\documentclass[letterpaper, 10 pt, conference]{Dconf}  

\IEEEoverridecommandlockouts                              

\usepackage{amsmath} 
\usepackage{amssymb}  
\usepackage{ascmac}
\usepackage{xcolor}
\usepackage[pdftex]{graphicx}
\usepackage{cite}
\usepackage{algorithmic}
\usepackage{algorithm}
\usepackage{multirow}

\title{\LARGE \bf
Language to Map: Topological map generation from natural language path instructions
}

\author{Hideki Deguchi$^{1}$, Kazuki Shibata$^{1}$ and Shun Taguchi$^{1}$
\thanks{$^{1}$They are work for Toyota central R\&D Labs., Inc., Yokomichi41-1, Nagakute, Aichi, Japan. © 2024 IEEE.  Personal use of this material is permitted.  Permission from IEEE must be obtained for all other uses, in any current or future media, including reprinting/republishing this material for advertising or promotional purposes, creating new collective works, for resale or redistribution to servers or lists, or reuse of any copyrighted component of this work in other works.}
}

\begin{document}

\maketitle
\thispagestyle{empty}
\pagestyle{empty}

\begin{abstract}
In this paper, a method for generating a map from path information described using natural language (textual path) is proposed. 
In recent years, robotics research mainly focus on vision-and-language navigation (VLN), a navigation task based on images and textual paths. 
Although VLN is expected to facilitate user instructions to robots, its current implementation requires users to explain the details of the path for each navigation session, which results in high explanation costs for users. 
To solve this problem, we proposed a method that creates a map as a topological map from a textual path and automatically creates a new path using this map. 
We believe that large language models (LLMs) can be used to understand textual path. 
Therefore, we propose and evaluate two methods, one for storing implicit maps in LLMs, and the other for generating explicit maps using LLMs. 
The implicit map is in the LLM’s memory. 
It is created using prompts. 
In the explicit map, a topological map composed of nodes and edges is constructed and the actions at each node are stored. 
This makes it possible to estimate the path and actions at waypoints on an undescribed path, if enough information is available. 
Experimental results on path instructions generated in a real environment demonstrate that generating explicit maps achieves significantly higher accuracy than storing implicit maps in the LLMs.
\end{abstract}

\section{INTRODUCTION}
In recent years, owing to the growing use of robots by people in their living environments, research on vision-and-language navigation (VLN) has accelerated\cite{vision, alfred}.
VLN is a navigation system that uses a camera image and natural language instructions to reach a set destination\cite{beyond, layout, vlmap, sasra}.
The use of natural language for explanations is expected to make it easier for users to provide instructions to robots. 

One of the problems with VLN is the high cost of instruction to explain the path detail.
To illustrate, instructions, such as ``Go straight out of the bedroom, turn right at the second corner and go down the stairs on the left...'', must be provided for each navigation. 
From the user's perspective, the ideal situation is to take to the destination simply by specifying its name.
While there are methods to search space by destination name\cite{building, ye2021auxiliary, chaplot2020object}, it would be more efficient to move to a destination if a path could be generated.

Therefore, a solution to this problem was proposed in this paper.
The proposed solution is to create a map as a topological map from textual path instructions. Then, users get path from created map only they instruct the destination name.
One way to create a map from natural language path instructions is to use large language models (LLMs)\cite{bert,llama,gpt3} with prompt\cite{systematic}.
However, although LLMs can facilitate the creation of implicit map through the use of an appropriate prompt, they have difficulties in processing long-term memory\cite{memorybank}, which may prevent them from recognizing the spatial structure and creating a correct map from the textual path. 

\begin{figure}[t]
    \centering
    \includegraphics[scale=0.555]{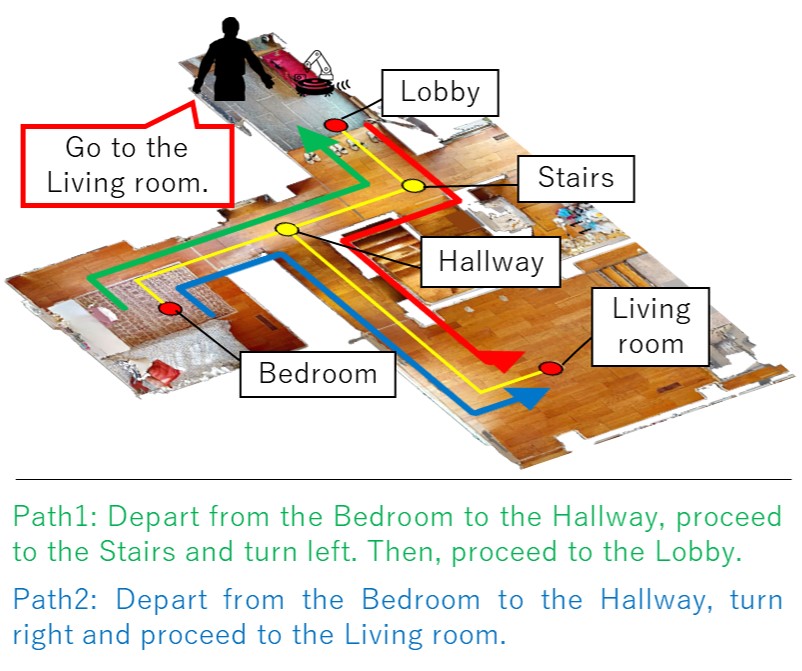}
    \caption{Image representation of task in this study. The above figure shows the map and the paths. The sentences in blue font underneath the figure describe the green and blue paths in the above figure. The task in this study was to input the detailed path as a bottom sentence and output the new path from the destination name.}
    \label{fig:intro}
\end{figure}

Then, we considered using LLMs to create an explicit map from textual maps for path description. 
The proposed explicit map was similar to a topological map \cite{toponav}. 
Our explicit map feature contained actions extracted from the textual path in each node. 
The stored actions could be used to output a path in the reverse direction, generate a new path, and so on. 
This frees the user from having to explain previously described paths or paths that can be inferred from those paths.
To the best of our knowledge, the proposed method is the first to map a space using solely natural language descriptions.

\begin{figure*}[t]
    \centering
    \includegraphics[scale=0.6]{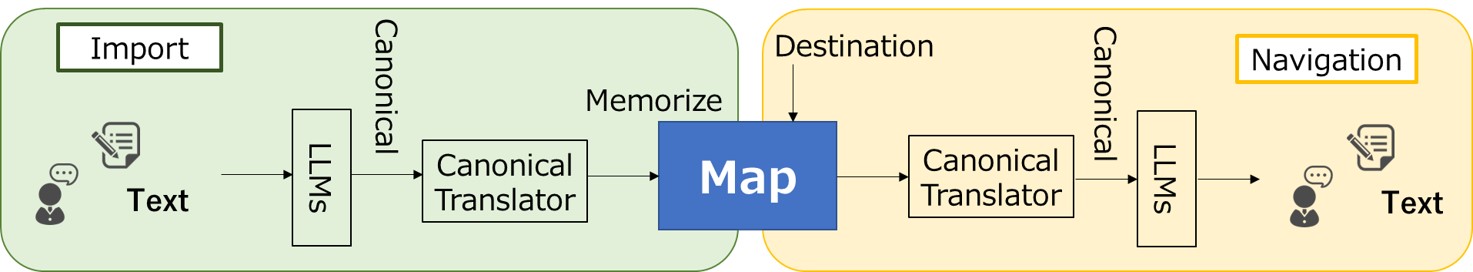}
    \caption{System overview. The left side shows the system that generates a map from user’s instruction. The right side shows the system that outputs the new path from destination name.}
    \label{fig:overview}
\end{figure*}

Figure \ref{fig:intro} shows an example task from this study. 
The sentences in the figure show the textual paths of the blue and green arrows in the environment.
The goal of this study was to create a map that, given the sentence in Figure \ref{fig:intro}, can output the path indicated by the red arrow above Figure \ref{fig:intro} from the name of the destination in natural language.

The contributions of this study are as follows:
\begin{itemize}
    \item Proposal of a method for memorizing textual path instructions and creating new ones that are not described.
    \item Proposal of a new map style that can be created from only natural language.
    \item Proposal of an algorithm for acquiring intermediate representations to convert natural languages into maps using LLMs.
    \item Confirmation by experiment that LLMs have poor spatial comprehension ability and that this ability can be improved by using explicit maps.
\end{itemize}

\section{RELATED WORK}
In recent years, there have been remarkable advances in LLMs that deal with natural language information\cite{bert, llama}. 
In particular, GPTs\cite{radford2018improving, radford2019better}, such as ChatGPT\cite{gpt3}, are being used in many areas, and research into their capabilities is ongoing\cite{summary}.
The use of LLMs is also advancing in the field of robotics, where they are being used for a variety of tasks\cite{data-e, saycan2022arxiv, palme, code}.

In the field of planning, there are studies that use natural language information to predict the scene of a destination or to determine the next action. 
Li et al.\cite{layout}  proposed a method for predicting the direction and scenery of a specified goal based on textual path instructions and 360-degree camera images using the results of learning various environmental data in advance.
Zehao et al.\cite{find} proposed a VLN method for path planning using geometric maps. 
Their method entailed comparing landmarks and actions in linguistic instructions and features of the geometric map to derive an appropriate path to the destination.
Zhou et al.\cite{navgpt} proposed a navigation method that used LLMs with a camera image to estimate the next action of a robot on its trajectory.
Shah et al.\cite{lm} proposed a method for planning a robot's path based on a topological map constructed from images and landmarks extracted from the textual path using LLMs.

In the field of mapping, natural language information processing by LLMs has been treated as auxiliary information for geometric mapping.
For example, in their study on VLN, Georagakis et al.\cite{cross} improved the success rate of navigation by predicting the blind spots in a camera image using path descriptions, in contrast to the conventional method of creating semantic maps from camera images.
Peihao et al.\cite{weakly} proposed a method for describing object information on semantic maps using the object information on paths in a textual path, thus enabling detailed object classification (e.g., color), beyond what was possible using existing segmentation models.
Chen et al.\cite{think} proposed a method that used language path instructions to generate topological maps in unknown environments. 
This method is similar to our proposed method, but Chen et al. differ significantly from our method in that they create a topological map from image information and then use linguistic information as supplementary information.

Thus, many scholars use LLMs for planning and mapping.
Most of them use pre-created geometric maps or camera images. 
In contrast, our method does not need to use them and creates a map using only natural language information.

\section{METHOD}\label{ch:method}
\subsection{System Overview}
To reduce the burden of path descriptions in VLN, we proposed a system that stores past path descriptions and automatically generates path descriptions when a new destination is obtained. 
In our system, multiple path descriptions are initially provided to the system.
The names of the starting points and destination are entered as queries at runtime. 
The system generates a map from the first input path descriptions, searches for a path on the internal map when a query is provided, converts the obtained path into a natural language description, and outputs the map.

In this study, we propose two methods; the first uses implicit maps and the other uses explicit maps. 
In the first method, all these inputs are given to the LLMs, and the map is stored in the LLMs. 
When a query is provided, LLMs generate a path based on an internally stored map. 
By contrast, in the method using explicit maps, the input is converted into a canonical representation by the LLMs. 
The map is then generated from the canonical information. 
When a query is given, pathfinding and action estimation are performed on the map, which is then converted into a natural language and output. 
The next section describes the use of the explicit map.

\subsection{Method using Explicit Map}

Since it is not certain whether LLMs can store implicit maps, this paper also proposes a method of having explicit maps, which will be compared and evaluated.
We show the overview of the proposed method using explicit map in Figure \ref{fig:overview}.

\subsubsection{Map Construction}
In this method, we construct topological map as a graph $G(N, E)$, consists of nodes $n \in N$ and edges $e \in E$.
We also store actions at each nodes $a_{n_i}(e_{ji}, e_{ik})$ defined as follows:
\begin{align}
\label{eq:action_definition}
a_{n_i}(e_{ji}, e_{ik}) &= \left\{\begin{array}{llrcl}
a_{\rm F} & \text{if} &-\theta &\leq \angle (e_{ji}, e_{ik}) &\leq \theta, \\
a_{\rm L} & \text{if} &\theta &< \angle (e_{ji}, e_{ik}) &< \pi - \theta, \\
a_{\rm R} & \text{if} &\pi + \theta &< \angle (e_{ji}, e_{ik}) &< - \theta, \\
a_{\rm T} & \text{if} &\pi - \theta &\leq \angle (e_{ji}, e_{ik}) &\leq \pi + \theta, \\
\end{array}
\right.
\end{align}
where $a_{n_i}(e_{ji}, e_{ik})$ denotes the action at the node $n_i$ through from the edge $e_{ji}$ to $e_{ik}$, $a_{\rm F}, a_{\rm L}, a_{\rm R}, a_{\rm T}$ represents the actions ``Forward'', ``Turn Left'', ``Turn Right'', ``Turn Around'', respectively.
As indicated by the above definition, each action defined in the language was divided into four parts based on the range of angles between the edges defined in the Euclidean space.
The $\theta$ is threshold of the angle.
The action of each node, which can be described in any linguistic expression, was approximated and assigned to one of four actions.

The proposed system uses canonical representation that can be translated from both language instructions and map when it translates using LLMs.
This is because it is known to be difficult to translate language instructions into formal information using LLMs, unlike vice versa\cite{pan2023data}.
In this study, canonical representation is a set of waypoints on the path $W$ and the actions at those waypoints.
For example of path 1 in Figure \ref{fig:intro}, the waypoints are 
\begin{align}
W =[{\rm Bedroom}, {\rm Hallway}, {\rm Stairs}, {\rm Lobby}],
\end{align}
and the actions are 
\begin{align}
a_{\rm Hallway}(e_{\rm Bedroom, Hallway}, e_{\rm Hallway, Stairs}) &= a_{\rm F}, \\
a_{\rm Stairs}(e_{\rm Hallway, Stairs}, e_{\rm Staris, Lobby}) &= a_{\rm L},
\end{align}
in canonical representation.
To construct a topological map, we added each waypoint as a node and each edge between waypoints as an edge and stored the actions at those waypoints.

\subsubsection{Path Finding and Instruction Generation}
To generate path instruction, we have to find the path between the start point and the destination of the query, and estimates action at each waypoints.

The path can be found on the topological map. 
One of the methods that can be used for finding the path is the use of the Dijkstra search algorithm.
However, it is necessary to estimate the action at each node from stored actions because they cannot be obtained by path finding on the topological map.
The target action, if present in the stored data, was used.
The target action, if not defined, was inferred from multiple actions, as follows:
\begin{align}
a_{n_i}(e_{ji}, e_{ik}) &= a_{n_i}(e_{ji}, e_{im}) a_{\rm T} a_{n_i}(e_{mi}, e_{ik}).
\end{align}
This is a transformation that uses the fact that the estimated action at node $n_i$ when moving from node, $n_j$, through $n_i$ to $n_k$ is equivalent to the action of moving from $n_j$ through $n_i$ to $n_m$, then turning around and going from $n_m$ through $n_i$ to $n_k$.

\begin{figure}[t]
    \centering
    \includegraphics[scale=0.4]{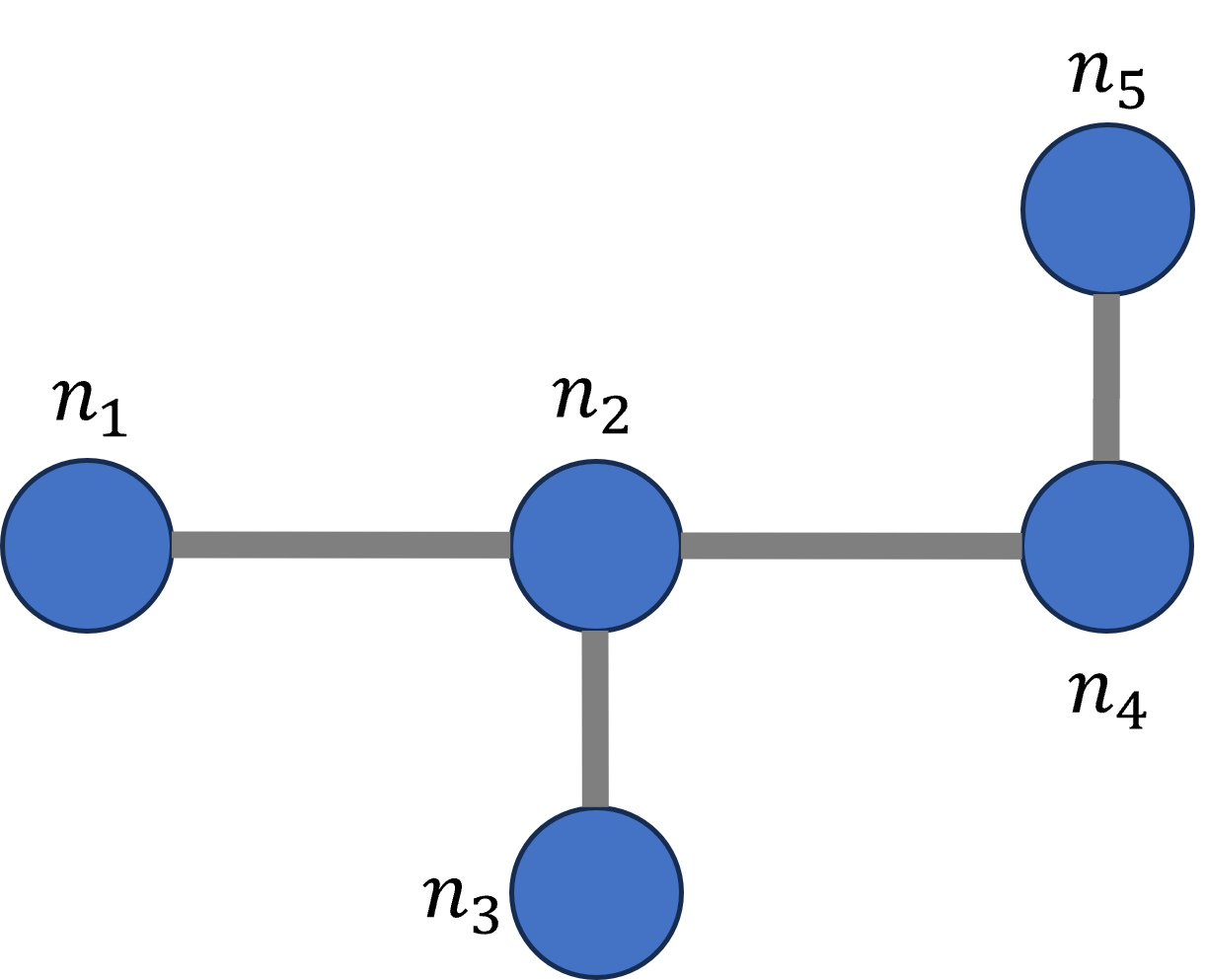}
    \caption{The environment is shown in the figure below, where $n_1$~$n_5$ is the name of each node.}
    \label{fig:toy}
\end{figure}

The following theorem, derived from \eqref{eq:action_definition} below, can be used to estimate the action.
\begin{align}
a_{n_i}(e_{ji}, e_{ik})^{-1} &= a_{n_i}(e_{ki}, e_{ij}), \\
a_{\rm F} a_{n_i}(e_{ji}, e_{ik}) &= a_{n_i}(e_{ji}, e_{ik}), \\
a_{n_i}(e_{ji}, e_{ik}) a_{\rm F} &= a_{n_i}(e_{ji}, e_{ik}), \\
a_{\rm L}^{-1} &= a_{\rm R}, \\
a_{\rm F}^{-1} &= a_{\rm F}, \\
a_{\rm T}^{-1} &= a_{\rm T}, \\
a_{\rm T} a_{\rm L} &= a_{\rm R}, \\
a_{\rm T} a_{\rm R} &= a_{\rm L}, \\
a_{\rm L} a_{\rm R} &= a_{\rm F}, \\
a_{\rm L} a_{\rm L} &= a_{\rm T}, \\
a_{\rm R} a_{\rm R} &= a_{\rm T}.
\end{align}

\begin{figure*}[t]
    \centering
    \includegraphics[scale=0.56]{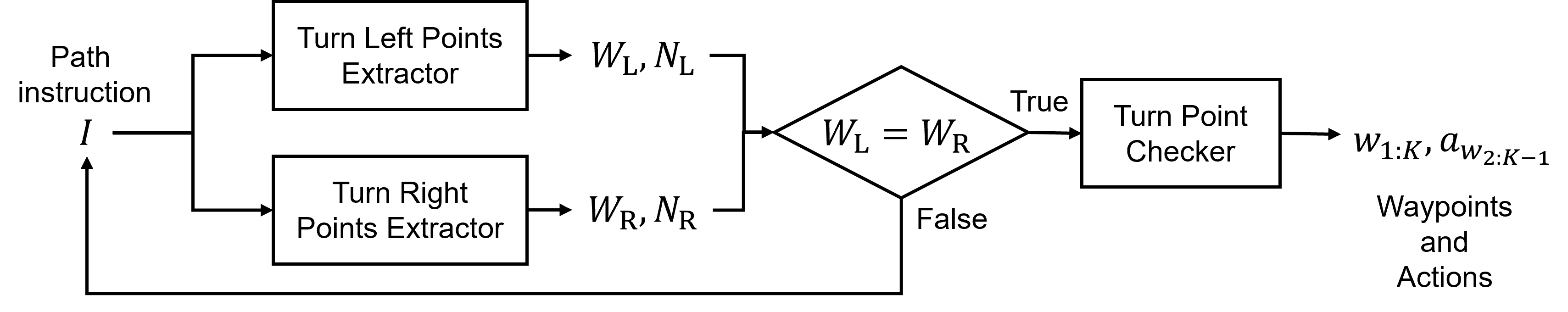}
    \caption{Flowchart of LLMs' prompt in this study.}
    \label{fig:prompt}
\end{figure*}

For clarity, we used the environment shown in Figure \ref{fig:toy}.
When user instructs the path of $n_1$ to $n_3$ and $n_1$ to $n_5$, our method memorizes these two path as below:
\begin{align}
W_1 &= [n_1, n_2, n_3], \\
W_2 &= [n_1, n_2, n_4, n_5].
\end{align}
Subsequently, the actions were also memorized, as follows:
\begin{align}
a_{n_2}(e_{12}, e_{23}) = a_{\rm R}, \\
a_{n_2}(e_{12}, e_{24}) = a_{\rm F}, \\
a_{n_4}(e_{24}, e_{45}) = a_{\rm L},
\end{align}
here, the path from $n_5$ to $n_3$ is $[n_5, n_4, n_2, n_3]$ can be estimated using path finding method.
The action of node $n_4$, and $n_2$ can be estimated as the following procedures:
\begin{align}
a_{n_4}(e_{54}, e_{42}) &= a_{n_4}(e_{24}, e_{45})^{-1} \nonumber\\
&= a_{\rm L}^{-1} \nonumber\\
&= a_{\rm R}, \\
a_{n_2}(e_{42}, e_{23}) &= a_{n_2}(e_{42}, e_{12}) a_{\rm T} a_{n_2}(e_{12}, e_{23}) \nonumber\\
&= a_{n_2}(e_{12}, e_{24})^{-1} a_{\rm T} a_{n_2}(e_{12}, e_{23}) \nonumber\\
&= a_{\rm F} a_{\rm T} a_{\rm R} \nonumber\\
&= a_{\rm L}. 
\end{align}
Finally we obtained, the following canonical representation:
\begin{align}
W = [n_5, n_4, n_2, n_3], \\
a_{n_4}(e_{54}, e_{42}) = a_{\rm R}, \\
a_{n_2}(e_{42}, e_{23}) = a_{\rm L}.
\end{align}
To output the path instruction, convert the canonical representations to natural language using LLMs.
Finally, we obtain the following path instruction: ``Depart from $n_5$ to $n_4$. Then, turn right and proceed to $n_2$. Then, turn left and proceed to $n_3$.''

\subsubsection{Translation to Canonical Representation using LLMs}
As explained above, the proposed method entailed translating natural language path instructions to canonical representations using LLMs. 
The extraction procedure for the canonical information is shown in Figure \ref{fig:prompt}.

First, path instructions input to LLMs using two different prompts.
Their extract the whole nodes in the path and left and right turn points respectively.
Our method is splitted into two parts, rather than prompting the LLM to output the left-turn points and right turn points simultaneously. It makes possible that reduces the difficulty of the task required for the LLM and increases the success rate.
Moreover, it also makes it possible to check for errors by comparing the output node sequences, thus improving the robustness of the system.
Specifically, if the output waypoints from each extractor are not equal, the process is repeated again.

Thenafter, the action of each waypoint is estimated.
If the waypoint is donoted as a turnpoint by one of the extractors, assign an action of $a_{\rm L}$ or $a_{\rm R}$ according to the extractor.
If neither, an action of $a_{\rm F}$ is assigned.
If the waypoint is denoted as a turnpoint by both extractors, the direction of rotation is determined by checking again with the LLM to determine the direction of rotation.

Complete algorithm is shown in Algorithm \ref{pr:translation}, where TurnLeftPointExtractor, TurnRightPointExtractor are the modules to extract waypoints and left turn (right turn) points.
TurnPointChecker is a module to check turn direction at point $w_k$.
$C$ is the maximum repetitive number or extractions.

\begin{algorithm}[t]
    \caption{Translation to Canonical Representation}
    \label{pr:translation}
    \begin{algorithmic}
        \STATE {\bf Input:} path instruction $I$
        \STATE {\bf Output:} waypoints $w_{1:K}$, actions $a_{w_{2:K-1}}$
        \STATE $W_{\rm L}, N_{\rm L} \leftarrow {\rm TurnLeftPointsExtractor}(I)$
        \STATE $W_{\rm R}, N_{\rm R} \leftarrow {\rm TurnRightPointsExtractor}(I)$
        \STATE $c = 0$
        \WHILE{$W_{\rm L} \neq W_{\rm R} \cap c < C$}
        \STATE $W_{\rm L}, N_{\rm L} \leftarrow {\rm TurnLeftPointsExtractor}(I)$
        \STATE $W_{\rm R}, N_{\rm R} \leftarrow {\rm TurnRightPointsExtractor}(I)$
        \STATE $c \leftarrow c + 1$
        \ENDWHILE
        \STATE $w_{1:K} \leftarrow W_{\rm R}$
        \FOR{$w_k \in \{w_{k} | w_{k} \in w_{2:K-1}\}$}
        \IF{$w_k \in N_{L} \cap w_k \in N_{R}$}
        \STATE $a_{w_k} \leftarrow {\rm TurnPointChecker}(I, w_k)$
        \ELSIF{$w_k \in N_{\rm L}$}
        \STATE{$a_{w_k} \leftarrow a_{\rm L}$}
        \ELSIF{$w_k \in N_{\rm R}$}
        \STATE{$a_{w_k} \leftarrow a_{\rm R}$}
        \ELSE
        \STATE{$a_{w_k} \leftarrow a_{\rm F}$}
        \ENDIF
        \ENDFOR
        \RETURN $w_{1:K}, a_{w_{2:K-1}}$
    \end{algorithmic}
\end{algorithm}

\section{EXPERIMENT}
\subsection{Dataset}
\label{sec:data}
\begin{figure}[ht]
    \centering
    \includegraphics[scale=0.35]{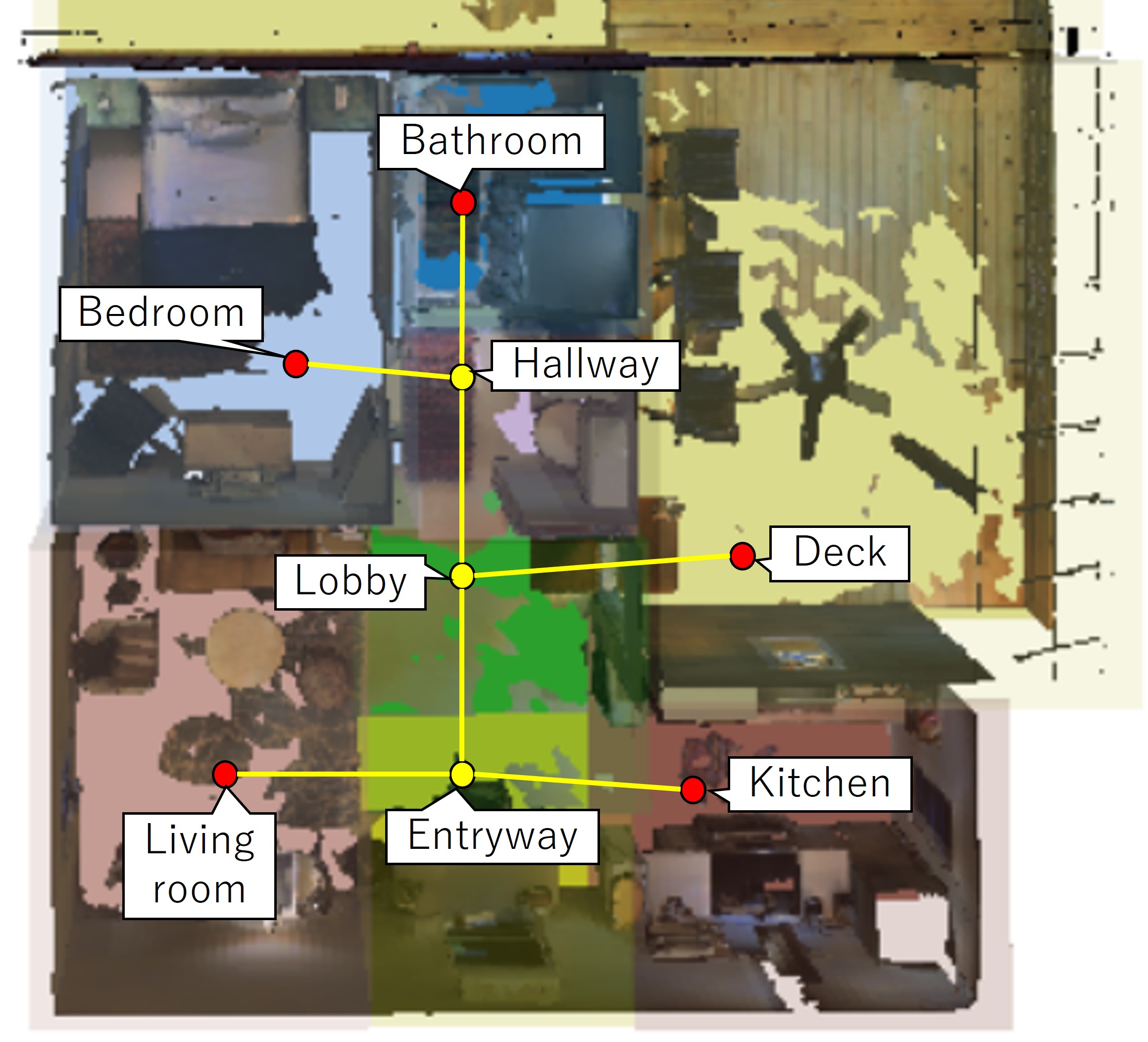}
    \caption{Example of the map with the path generation. The red circles describe the start or destination nodes and the yellow circles describe the waypoint nodes. The yellow lines show the edge.}
    \label{fig:jh}
\end{figure}

In this study, the effectiveness of implicit and explicit map storage methods for inferring spatial structure from path instructions using LLMs was evaluated.
Graph maps were manually created for models in the Matterport3D dataset\cite{matterport3d} and used as the evaluation environment.
The path instructions were generated manually for these environments.
Figure \ref{fig:jh} shows an example of the environment used.
We selected 10 environments and five nodes as the start or destination nodes in each environment. 
We then created 10 textual paths by selecting two pairs from these five nodes.
In this study, we assumed the following for the user path description:
\begin{itemize}
    \item The user has a name at each node of the topological map in mind.
    \item The paths are created such that they have the minimum nodes from start to destination.
    \item The user doesn't describe landmarks not related to the action.
    \item The user describe all the landmarks which related to the action in their path description.
    \item The user doesn't miss their instructions.
\end{itemize}

\subsection{Evaluation}
In this experiment, the following two evaluations were conducted:
\begin{description}
    \item[Reverse path]~\\
    Create the reverse path of the input path.
    \item[Combined path]~\\
    9 paths were selected and input from the 10 paths prepared for each environment. One excluded path was generated from the information on the start and goal names of the paths.
\end{description}
The reverse path was evaluated to assess the system’s ability to grasp the spatial structure from the path input. 
The combined path was evaluated to assess the ability of the system to generate a new path from multiple input paths.
When evaluating the combined path, we use two evaluation metrics, reachable path and shortest path. Reachable path allows the path not to be the shortest. Note that  the shortest route here means the route with the smallest number of edges on the topological map. 
Perform the experiment with the 100 paths described in Section \ref{sec:data} and derive the success rate.

\begin{table}[t]
    \centering
    \caption{Prompts used in experiment.}
    \begin{tabular}{p{1.0cm}|p{1.5cm}|p{5cm}}
        \multicolumn{2}{c|}{Types of Prompt} & Prompt \\ \hline \hline
        Implicit method & Reverse path & Show the reverse path, reversing the start and goal of the following path. \\ \cline{2-3}
        & Combined path & Understand the spatial structure of path1-9 below and create the shortest path from the specified start to goal. However, be sure to indicate the action to be taken at each passing point. \\ \hline
        Explicit method & Turn points extractor & Extract waypoints in the description of the navigation path. Then, extract the points which turn left/right. \\ \cline{2-3}
        & Turn points checker & For the following path, answer the action at specified point is turn right or left. \\ \hline
    \end{tabular}
    \label{tab:prompt}
\end{table}

\subsection{Setting}
Table \ref{tab:prompt} shows the prompts which use our method in the experiment.
Implicit method for reverse path and combined path in Table \ref{tab:prompt} are simple prompts to evaluation.
Turn points extractor and Turn points checker in Table \ref{tab:prompt} are the prompts at explicit map method.
The turn points extractor is purposely implemented in two steps: the extraction of all waypoints and the extraction of left/right turn points.
In this study, we use OpenAI API\cite{gpt3.5,gpt4} and the GPT function calling feature is used in the implementation.


\subsection{Results}
Tables \ref{tab:inv_result} and \ref{tab:9to1_result} present the results of the representation reverse path and combined path using LLMs evaluation.
Implicit/Explicit method in tables indicates the method that uses an implicit/explicit map.

The result of Table \ref{tab:inv_result} shows a 60\% success rate for the simple task of deriving a path in the opposite direction of the described path.
However, the success rate of the proposed method using explicit maps and GPT-4\cite{gpt4} was $>$ 90\% for LLMs.
Therefore, the proposed method with explicit map almost solved the inverse generation problem.

Based on the result of Table \ref{tab:9to1_result}, the proposed method using an implicit map could not create new paths from textual paths.
However, the proposed method using an explicit map effectively clarified the spatial structure of paths described in natural language and generated new paths. 
The failed cases were errors in extracting waypoints and actions from natural language (Turn left/right extractor modules in fig.\ref{fig:prompt} ).

An example of the input/output using the environment and path sentences shown in Figure \ref{fig:jh} is shown in Figure \ref{fig:paths}.
Verifying the results of Figure \ref{fig:paths} against Figure \ref{fig:jh} reveals that the proposed method using an implicit map output the wrong direction for the entry way turn, whereas the proposed method using an explicit map output the correct path to the destination.
The Figure \ref{fig:variety_paths} shows the results of a qualitative evaluation conducted using LLMs\cite{gpt4} with various representations of textual paths as input.
The result of Figure \ref{fig:variety_paths} indicates the explicit method can create the map and the new path from various representations.

\begin{table}[t]
    \centering
    \caption{SUCCESS RATE OF REVERSE PATHS EVALUATION.}
    \begin{tabular}{ll|c}
        Method          & LLM            & Success rate \\ \hline \hline
        Implicit method & GPT-3.5-turbo  & 67\% \\
                        & GPT-4          & 66\% \\ \hline
        Explicit method & GPT-3.5-turbo  & 83\% \\
                        & GPT-4          & \bf{94\%} \\ \hline
    \end{tabular}
    \label{tab:inv_result}
\end{table}

\begin{table}[t]
    \centering
    \caption{SUCCESS RATE OF COMBINED PATHS EVALUATION.}
    \begin{tabular}{ll|c|c}
        Method & LLM & \multicolumn{2}{c}{Success rate} \\
        & & Reachable path & Shortest path \\ \hline \hline
        Implicit method  & GPT-3.5-turbo  & 19\% & 8\% \\
                         & GPT-4          & 44\% & 10\% \\ \hline
        Explicit method  & GPT-3.5-turbo  & 77\% & 77\% \\
                         & GPT-4          & \bf{92\%} & \bf{92\%} \\ \hline
    \end{tabular}
    \label{tab:9to1_result}
\end{table}

\section{DISCUSSION}
The results in Table \ref{tab:9to1_result} show that the existing LLMs failed approximately 40\%. 
One of the most common causes of LLMs failure is incorrect rotational position or direction in the path(96\%). 
This suggests that LLMs lack the ability to grasp the spatial structures of language paths.
This becomes clearer when we consider the evaluation results of the combined path. 
In the combined path evaluation, methods using only LLMs were almost unable to show the shortest path, while the method using only GPT-4 showed a success rate of about 40\% for the reachable path.
Therefore, it can be seen that LLMs have the ability to use the given path sentences as they are but do not have the ability to recombine them to generate appropriate paths. 
By contrast, the proposed method using an explicit map had a high success rate based on both evaluations and fully compensated for the weakness of LLMs in understanding the spatial structure.
The method that uses an explicit map can always output the correct path if the intermediate representation is accurate.
Hence, the accuracy can be improved by making the prompt to extract canonical information from the textual path more accurate.

\begin{figure}[t]
    \begin{screen}
        \begin{description}
            \item[Start point: ``Living room'']
            \item[Goal point: ``Bathroom'']
        \end{description}
        \begin{description}
            \item[\textcolor{teal}{Path 1}] \textcolor{teal}{Depart from the "Living room" to the "Entryway", and proceed to the "Kitchen".}
            \item[\textcolor{teal}{Path 2}] \textcolor{teal}{Depart from the "Kitchen" to the "Entryway", turn right and proceed to the "Lobby". Continue on to the "Hallway" and proceed to the "Bathroom".}
        \end{description}
        \begin{description}
            \item[\textcolor{blue}{Implicit method}]~\\
            \textcolor{blue}{From the "Living Room", proceed to the "Entryway" (Path 1).\\
From the "Entryway", turn right and head to the "Lobby" (Path 2).\\
From the "Lobby", continue to the "Hallway" (Path 2).\\
From the "Hallway", proceed to the "Bathroom" (Path 2).}
        \end{description}
        \begin{description}
            \item[\textcolor{red}{Explicit method}]~\\
            \textcolor{red}{Depart from the living room and advance to the entryway, then turn left. From there, advance to the lobby and continue on to the hallway, then advance further to the bathroom where you will arrive.}
        \end{description}
    \end{screen}
    \caption{An example of the textual path in Figure \ref{fig:jh} used for the quantitative evaluation. \textcolor{teal}{The green sentences show the textual paths}, \textcolor{blue}{the blue sentence shows the output of the implicit map method} and \textcolor{red}{the red sentence shows the output of the explicit map method}. This figure shows only the paths used for output from the 9 input paths. The notation ``(Path 1)'' or ``(Path 2)'' is itself contained in the raw output of the implicit method. These phrases indicate which textual paths were used to create the new path.}
    \label{fig:paths}
\end{figure}

\begin{figure}[t]
    \begin{screen}
        \begin{description}
            \item[Start point: ``Deck'']
            \item[Goal point: ``Kitchen'']
        \end{description}
        \begin{description}
            \item[\textcolor{teal}{Path 1}] \textcolor{teal}{Commence on the ``Deck'', walk to the "Lobby", swing left there and towards the "Entryway". Then, guide yourself right there and go to the ``Living room''.}
            \item[\textcolor{teal}{Path 2}] \textcolor{teal}{Begin in the ``Living room'', move through the "Entryway", and then head straight to the "Kitchen".}
            \item[\textcolor{blue}{Implicit method}]~\\ \textcolor{blue}{1. Start at the "Deck", go to the "Lobby" (Path 1). \\
2. From the "Lobby", go to the "Entryway" (Path 1). \\
3. From the "Entryway", go straight to the "Kitchen" (Path 2).}
            \item[\textcolor{red}{Explicit method}]~\\
            \textcolor{red}{Depart from the deck and advance to the lobby, then turn left. From the lobby, continue on to the entryway, again turning left. From there, advance to the kitchen where you'll arrive.}
        \end{description}
    \end{screen}
    \caption{Paths converted with LLMs\cite{gpt4} to more diverse wording than the input path of Figure \ref{fig:jh}. The explicit method can output path descriptions more accurately than the implicit method, even if the expression of the instructions changes.}
    \label{fig:variety_paths}
\end{figure}

The main limitation of this study is the assumption regarding the inputs. 
First, the proposed method does not deal with the diversity of instruction style. For example, the proposed method requires to describe path information from one starting point to another. However, users may describe the locations of multiple points from a single starting point in one time. 
In this case, we need to make a system that decomposes the instruction into multiple paths and it is future work.

Second, this study assumes that users do not miss instructions. However, user instructions sometimes contain errors, against which the proposed method must prove robust when used in a real environment. To address this issue, we plan to introduce a system that detects discrepancies in explanations from the generated explicit maps and asks the user whether the instructions are correct.

Finally, we evaluated a small environment, similar in size to a house, because this was sufficient to compare the implicit and explicit map methods. However, there are many large environments, such as shopping malls, which our explicit map method must address. In future, we will evaluate how large a space can be mapped using our proposed method.

\section{CONCLUSION}
In this study, we focused on the task of creating a map from a path described in a natural language. To address this issue, we proposed two methods, one that implicitly memorize the paths in LLMs and another that explicitly create maps. The implicit map method uses the prompt and stores the spatial structure of the LLMs implicitly. The explicit map method uses a prompt to create a map that explicitly consists of waypoints and actions at each waypoint of the textual path. The experiments were conducted using paths generated on a graph map created in a real environment. The experimental results show that existing LLMs have a success rate of only approximately 65\% for tasks that output the reverse of a given path, whereas the method using the map has a success rate of over 90\%, even for difficult recombination tasks. In the future, we plan to improve the proposed method so that it can handle ambiguous user instructions and descriptions in various spaces, and extend our method to a path planning that can consider the dynamics of the robot.





\clearpage

\bibliographystyle{Dtran}
\bibliography{deguchi_arXiv}             

\end{document}